\documentclass{SCIS2026}

\usepackage{array}
\usepackage{paralist}
\usepackage{csquotes}
\usepackage{cleveref}
\usepackage[section]{placeins}

\newcommand{\etc}{etc.}
\newcommand{\etal}{et al.}

\makeatletter
\providecommand{\headerps@out}[1]{}
\makeatother

\begin{document}

\ArticleType{RESEARCH PAPER}
\Year{2025}
\Month{January}
\Vol{68}
\No{1}
\DOI{}
\ArtNo{}
\ReceiveDate{}
\ReviseDate{}
\AcceptDate{}
\OnlineDate{}
\AuthorMark{Mao Y et al.}
\AuthorCitation{Mao Y, Qin Z, Zhou J, et al. Autoregressive image generation with linear complexity: a spatial-aware decay perspective}

\title{Autoregressive Image Generation with Linear Complexity: A Spatial-Aware Decay Perspective}{Autoregressive image generation with linear complexity}
\author[1]{Yuxin Mao}{}
\author[2]{Zhen Qin}{}
\author[3]{Jinxing Zhou}{}
\author[1]{Hui Deng}{}
\author[2]{Xuyang Shen}{}
\author[1]{\\[-1mm]Bin Fan}{}
\author[4]{Jing Zhang}{}
\author[5]{Yiran Zhong}{}
\author[1]{Yuchao Dai}{{daiyuchao@gmail.com}}

\address[1]{School of Electronics and Information, Northwestern Polytechnical University, Xi'an, China}
\address[2]{Seed, ByteDance, Shanghai, China}
\address[3]{OpenNLPLab, Shanghai, China}
\address[4]{The Australian National University, Canberra, Australia}
\address[5]{Tongyi Laboratory, Alibaba, Shanghai, China}

\abstract{Autoregressive (AR) models have recently emerged as a promising paradigm for image generation, as they can model visual tokens through the same next-token prediction principle that underlies large language models. However, most competitive AR image generators still rely on Transformer attention, whose quadratic token interaction cost and growing key-value cache become increasingly expensive for high-resolution visual sequences. Linear attention provides an attractive route to efficient AR generation, but our study shows that directly importing linear or decay-based attention from language modeling leads to clear quality degradation. We identify a key cause: after a 2D token map is flattened by raster scan, 1D sequence locality no longer faithfully reflects spatial locality, and conventional decay mechanisms may introduce a false-locality bias at row transitions. To address this problem, we propose Linear Attention with Spatial-Aware Decay (LASAD), a simple recurrent attention mechanism that modulates decay according to the original 2D token geometry while preserving linear complexity. Built upon this layer, LASADGen replaces standard attention in an AR image generator with spatial-aware linear attention. Experiments on ImageNet demonstrate that LASADGen achieves a favorable quality-efficiency tradeoff, improving generation quality over linear-attention baselines while reducing inference cost and memory usage compared with Transformer-based AR models.}


\maketitle

\section{Introduction}
\label{sec:intro}

Deep generative modeling has made rapid progress in image synthesis, driven by advances in GANs~\cite{biggan,gigagan,stylegan-xl}, diffusion models~\cite{dhariwal_adm_nips_2021diffusion,cdm,ldm,dit,ho_ddpm_nips_2020}, and, more recently, autoregressive (AR) visual generators~\cite{llamagen,var,mao2024tavgbench,qin_lightnet_2024}. 
Among these paradigms, AR image generation is particularly appealing because it formulates visual synthesis as sequential token prediction, closely mirroring the training objective that has enabled the success of large language models or multimodal models~\cite{touvron2023llama,radford2018improving,liu2024deepseekv3,mao2023multimodal,mao2024generative,mao2025contrastive,hao2024improving}. 
By representing an image as a sequence of discrete visual tokens, AR models can capture local texture, object-level layout, and long-range compositional structure within a unified likelihood-based framework. 
This compatibility with language-model architectures also makes AR modeling a promising route toward unified visual and multimodal generation.

Despite these advantages, efficient visual AR generation remains challenging. 
Most competitive AR image generators adopt Transformer attention~\cite{vaswani2017attention}, where each token interacts with all preceding tokens. 
For a visual token sequence of length $N$, this design incurs quadratic $O(N^2)$ attention cost during training and requires a growing key-value (KV) cache during autoregressive inference. 
These costs become especially restrictive when the image resolution increases, the tokenizer down-sampling ratio decreases, or the model is extended to longer visual sequences. 
In language modeling, linear attention and related recurrent or hybrid architectures have been widely explored to reduce this burden~\cite{shen2024scaling,qin_tnl_various,li2025minimax,lieber2024jamba,mao2026learning,mao2026ldt}. 
This naturally raises the central question of this work: \textit{Can linear attention be used for autoregressive image generation without sacrificing generation quality?}

Our empirical study suggests that the answer is not straightforward. 
When we directly replace the multi-head self-attention layers in a strong AR image generator with ordinary linear attention, generation quality drops substantially compared with the SoftmaxAttn baseline (see~Table~\ref{tab:abl_arch_eval}). 
This degradation is not fully resolved by adopting recent decay-based linear attention mechanisms from language modeling~\cite{yang_GLA_2023,sun2023retentive,qin_hgrn_nips_2024,qin2024hgrn2,chen2025minimaxM1}. 
Although such decay mechanisms improve sequence modeling by emphasizing nearby tokens in a 1D order, they are not designed for the geometry of images. 
As a result, the standard linear-attention recipe that works well for natural language does not directly transfer to raster-scanned visual tokens.

We argue that this failure stems from a structural mismatch between 1D sequence locality and 2D image locality. 
Images are inherently two-dimensional, but AR image generators commonly flatten the token map into a 1D sequence using raster scan. 
This operation preserves the left-to-right order within each row, but it also creates false neighbors at row transitions: the last token of one row and the first token of the next row are adjacent in the flattened sequence, even though they are far apart in the original 2D grid. 
Conventional 1D decay mechanisms treat these tokens as maximally local because their sequence distance is one. 
This injects an inappropriate inductive bias into the recurrent attention state, encouraging information flow that is local in sequence order but not local in image space. 
For visual generation, where spatial coherence is crucial, such false locality can damage the model's ability to preserve authentic 2D structure.

To address this issue, we propose Linear Attention with Spatial-Aware Decay (LASAD), a simple yet effective attention mechanism tailored for raster-scanned visual sequences. 
The key idea is to make the decay process aware of the original token-grid geometry rather than relying solely on 1D token positions. 
Specifically, LASAD modifies the decay behavior at row-boundary positions, allowing the recurrent state update to distinguish true spatial neighbors from artificial neighbors introduced by flattening. 
This design retains the efficiency of recurrent linear attention while injecting a minimal spatial prior that is aligned with the 2D structure of images. 
We further build LASADGen by replacing conventional attention layers in the LlamaGen framework~\cite{llamagen} with LASAD layers, thereby preserving the AR generation pipeline while improving its quality-efficiency tradeoff.

We evaluate LASADGen on class-conditional ImageNet generation~\cite{deng2009imagenet}. 
Across model scales, LASADGen consistently improves over ordinary linear attention and decay-based linear attention baselines, and it achieves competitive or better generation quality compared with Transformer-based AR counterparts under comparable settings. 
Meanwhile, the linear recurrent formulation reduces inference time and memory usage by avoiding the growth of a Transformer-style KV cache. 
These results indicate that the key obstacle to efficient AR image generation is not linear attention itself, but the lack of a spatially compatible decay mechanism for flattened visual tokens.

Our contributions are summarized as follows:
\begin{compactitem}
\item We identify the false-locality problem that arises when 1D decay-based linear attention is applied to raster-scanned image token sequences, showing that sequence adjacency does not always imply spatial adjacency.

\item We propose LASAD, a spatial-aware linear attention mechanism that modulates recurrent decay according to the original 2D token geometry while maintaining linear complexity.

\item We develop LASADGen, an efficient autoregressive image generator based on LASAD layers, and demonstrate its favorable quality-efficiency tradeoff through comprehensive ImageNet experiments.
\end{compactitem}

\section{Related Works}
\label{sec:related_works}

\noindent\textbf{Autoregressive Visual Generation Methods.}
Visual generation has gained significant attention in recent years, particularly with the advancement of deep learning architectures and generative models~\cite{dit, ldm, llamagen,mao2024tavgbench}. 
A key trend is the use of autoregressive models~\cite{touvron2023llama, radford2018improving}, which exhibit substantial generality and potential, owing to their strong connection with natural language processing (NLP). 
Autoregressive generation methods are typically categorized into two main approaches: BERT-style masked autoregressive models and GPT-style autoregressive models. 
Masked autoregressive methods~\cite{chang2022maskgit,li_Mage_cvpr_2023,li_MAR_nips_2025, yu2023magvit}, inspired by BERT-style pretraining~\cite{devlin2019bert}, generate images by predicting randomly masked tokens. 
In contrast, GPT-style autoregressive models~\cite{esser2021taming, yu2021vector, radford2018improving} employ image tokenization techniques~\cite{kingma2013auto, van2017neural} to convert images into discrete representations, enabling sequential next-token prediction.
Recently, LlamaGen~\cite{llamagen} extends autoregressive methods by adapting Llama~\cite{touvron2023llama} and similar large language model architectures for image token prediction.
By applying the \enquote{next-token prediction} paradigm to visual generation, LlamaGen demonstrates competitive performance in image generation tasks. 
Meanwhile, VAR~\cite{var} adjusts the paradigm of autoregressive generation and design \enquote{next-scale prediction}, which is also a successful application of autoregression in visual generation tasks.
Due to the scalability of LlamaGen, there are still many works expanding on it, such as visual token decoding methods~\cite{wang2024parallelized,pang2024next,he2024zipar,yu2025randomized}, stronger visual tokenizers~\cite{li2024imagefolder}, more effective model structures~\cite{li2024scalable,liu2024elucidating}, \etc.

\noindent\textbf{Linear Attention.} Linearized attention challenges the traditional quadratic complexity of softmax attention by utilizing kernel-based methods to decompose attention computations. Rather than computing the full attention matrix, linear attention projects queries and keys into a transformed space, where the interactions can be computed with linear complexity $O(N)$ relative to the sequence length. 
The evolution of linear attention has followed several innovative trajectories. 
Katharopoulos~\etal~\cite{katharopoulos_transformers_are_rnn_ICML_2020} posits that positive kernel functions, such as $elu(x)+1$, can be used to approximate nonlinear \textbf{softmax} function. Consequently, this enables linearization of attention while maintaining numerical stability.
This insight paved the way for a new family of attention variants. 
The Random Feature Attention (RFA) framework~\cite{peng2020random} introduced probabilistic approximations using random Fourier features, while Performer~\cite{choromanski2020rethinking} advanced this concept by designing positive random features that better preserve the exponential characteristics of softmax attention. 
Cosformer~\cite{qin_iclr_cosformer} decomposes attention based on trigonometric principles, achieving linear complexity while preserving crucial theoretical properties of softmax attention. 
Recent advances in linear attention mechanisms frequently incorporate decay functions to enhance token locality, as demonstrated in models such as TNL~\cite{qin_tnl_various}, GLA~\cite{yang_GLA_2023}, RetNet~\cite{sun2023retentive}, and HGRN~\cite{qin_hgrn_nips_2024, qin2024hgrn2}, \etc. 
This decay approach effectively addresses the uniform weighting limitation inherent in traditional linear attention by exponentially discounting contributions from older tokens, preventing attention saturation issues.


\noindent\textbf{Uniqueness of Our Solutions.}
Although linear attention has been extensively explored and applied across many domains, our approach goes beyond the naive application of linear attention to autoregressive image generation by addressing a fundamental limitation: the dimensional mismatch that arises when inherently 2D image data are flattened into 1D token sequences. Such flattening disrupts the spatial relationships among image patches, which are essential for maintaining structural coherence during image generation. 
Consequently, while linear attention provides clear computational efficiency advantages, it remains critical to investigate mechanisms that can effectively preserve spatial context in visual data. 
Our study focuses on the interaction between sequence modeling and spatial structure, and is therefore orthogonal to the specific image representation schemes adopted by existing autoregressive image generation methods.
\section{Method}
\label{sec:method}

This section presents LASADGen, an autoregressive image generator built with Linear Attention with Spatial-Aware Decay (LASAD). 
We first formulate visual autoregressive generation on raster-scanned image tokens. 
We then revisit linear attention and decay-based linear attention, explaining why a decay rule designed for 1D language sequences is not directly suitable for flattened 2D image tokens. 
Finally, we introduce the proposed spatial-aware decay mechanism, describe the resulting LASADGen architecture, and analyze its computational properties.

\subsection{Autoregressive Visual Generation}
\label{sec:sec3.1}

Given an image $I \in \mathbb{R}^{H \times W \times 3}$, an image tokenizer first maps it into a discrete token grid $X \in \mathbb{R}^{h \times w}$, where $h=H/p$ and $w=W/p$ for tokenizer down-sampling ratio $p$. 
Following standard autoregressive visual generators~\cite{llamagen}, the 2D token grid is flattened into a 1D sequence by raster scan:
\begin{equation}
X \mapsto \mathbf{x}=(x_1,x_2,\ldots,x_N), \qquad N=h\times w.
\end{equation}
The token index $t$ corresponds to a 2D coordinate
\begin{equation}
r(t)=\left\lfloor\frac{t-1}{w}\right\rfloor,\qquad
c(t)=(t-1)\bmod w,
\end{equation}
where $r(t)$ and $c(t)$ denote the row and column indices in the original token grid.

Autoregressive image generation factorizes the likelihood as sequential next-token prediction:
\begin{equation}
\label{eq:ar_model}
p(\mathbf{x})=\prod_{t=1}^{N}p(x_t|x_1,x_2,\ldots,x_{t-1})
=\prod_{t=1}^{N}p(x_t|x_{<t}).
\end{equation}
For class-conditional generation, a conditioning token or embedding $c$ is prepended to the visual sequence, giving
\begin{equation}
p(\mathbf{x}|c)=\prod_{t=1}^{N}p(x_t|x_{<t},c).
\end{equation}
The model is trained with the standard next-token cross-entropy objective:
\begin{equation}
\mathcal{L}_{\text{train}}
=\text{CE}\left(\mathbf{M}([c,x_1,x_2,\ldots,x_{t-1}]),[x_1,x_2,\ldots,x_t]\right),
\end{equation}
where $\mathbf{M}$ denotes the autoregressive generator. 
In Transformer-based AR generators, $\mathbf{M}$ usually relies on causal self-attention, whose token interaction cost grows quadratically with $N$. 
This motivates replacing softmax attention with a more efficient recurrent attention mechanism.

\subsection{Linear Attention and Decay}
\label{sec:sec3.2}

\noindent\textbf{Softmax attention.}
For query, key, and value matrices $\mathbf{Q},\mathbf{K},\mathbf{V}\in\mathbb{R}^{N\times d}$, standard attention computes
\begin{equation}
\label{equ:softmax_atten}
\text{Attention}(\mathbf{Q},\mathbf{K},\mathbf{V})
=\text{softmax}\left(\frac{\mathbf{Q}\mathbf{K}^{\top}}{\sqrt{d}}\right)\mathbf{V}.
\end{equation}
The explicit attention matrix has $O(N^2)$ time and memory complexity. 
During autoregressive inference, Transformer attention also stores a KV cache whose size grows linearly with sequence length and layer depth.

\noindent\textbf{Linear attention.}
Linear attention replaces the softmax kernel with a feature map $\phi(\cdot)$ and reorders the matrix product~\cite{katharopoulos_transformers_are_rnn_ICML_2020,qin_iclr_cosformer,choromanski2020rethinking}:
\begin{equation}
\mathbf{O}
=\mathbf{\Delta}^{-1}\phi(\mathbf{Q})[\phi(\mathbf{K})^{\top}\mathbf{V}],\qquad
\mathbf{\Delta}=\mathrm{diag}\big(\phi(\mathbf{Q})[\phi(\mathbf{K})^{\top}\mathbf{1}_{N}]\big).
\end{equation}
Computing $\phi(\mathbf{K})^{\top}\mathbf{V}$ before multiplying by $\phi(\mathbf{Q})$ avoids forming the full attention matrix. 
Following prior work~\cite{qin_transnormer_emnlp_2022}, the denominator can also be replaced by a normalization layer:
\begin{equation}
\label{linear_attention}
\mathbf{O}=\mathrm{Norm}\left(\phi(\mathbf{Q})[\phi(\mathbf{K})^\top\mathbf{V}]\right).
\end{equation}
In the causal setting, linear attention admits a recurrent form:
\begin{equation}
\begin{gathered}
\mathbf{s}_0=\mathbf{0},\qquad
\mathbf{s}_t=\mathbf{s}_{t-1}+\mathbf{k}_t\mathbf{v}_t^\top,\qquad
\mathbf{o}_t^\top=\mathbf{q}_t^\top\mathbf{s}_t,\\
\mathbf{q}_t,\mathbf{k}_t,\mathbf{v}_t,\mathbf{o}_t\in\mathbb{R}^{d},\qquad
\mathbf{s}_t\in\mathbb{R}^{d\times d},\qquad t=1,\ldots,N.
\end{gathered}
\end{equation}
This recurrence can be evaluated with linear complexity in sequence length. 
However, as shown in~Table~\ref{tab:abl_arch_eval}, directly replacing softmax attention with ordinary linear attention substantially degrades image generation quality.

\begin{figure}[t]
  \centering
  \setlength{\abovecaptionskip}{0.2cm}
    \includegraphics[width=0.45\linewidth]{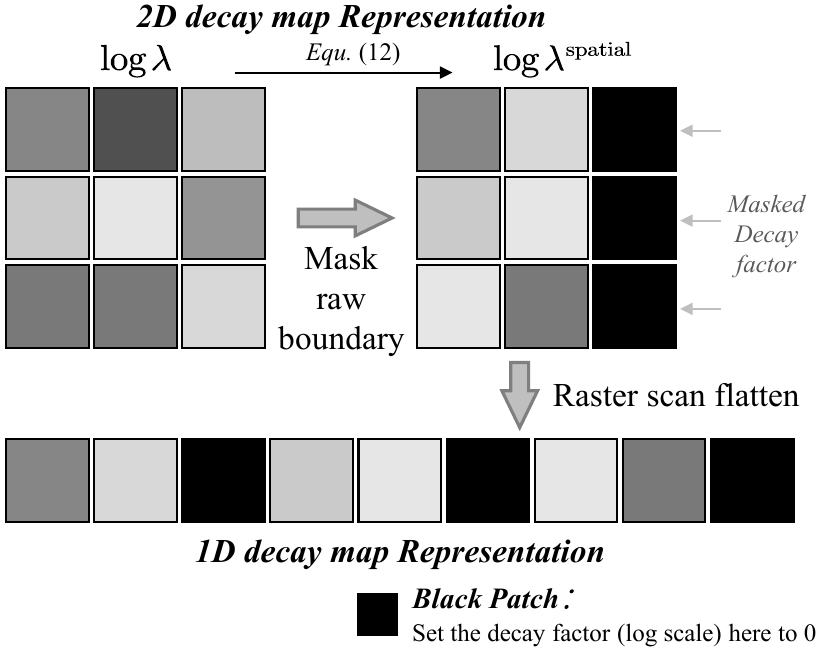}
    \caption{\textbf{Visualization of the Spatial-Aware Decay mechanism}. 
    Each color block represents the decay factor of the token at the corresponding position.
    Our method masks row-boundary positions when raster-scan flattening creates artificial sequence neighbors, preventing the standard 1D decay rule from treating non-adjacent image tokens as ordinary local neighbors.}
    \label{fig:decay_mask}
\end{figure}

\subsection{Why 1D Decay Fails for Flattened Images}
\label{sec:sec3.3}

Recent efficient sequence models improve linear attention by introducing a decay factor into the recurrent state~\cite{2307.14995,qin_hgrn_nips_2024,qin2024hgrn2}. 
The generic form is
\begin{equation}
\begin{gathered}
\mathbf{s}_0=\mathbf{0},\qquad
\mathbf{s}_t=\mathrm{diag}(\lambda_t)\mathbf{s}_{t-1}
+\mathbf{k}_t\mathbf{v}_t^\top,\qquad
\mathbf{o}_t^\top=\mathbf{q}_t^\top\mathbf{s}_t,\\
\mathbf{q}_t,\mathbf{k}_t,\mathbf{v}_t,\mathbf{o}_t,\lambda_t\in\mathbb{R}^{d},\qquad
\mathbf{s}_t\in\mathbb{R}^{d\times d},\qquad
0\leq\lambda_t\leq1.
\end{gathered}
\end{equation}
The decay term controls how much historical information is retained when the state moves from position $t-1$ to $t$. 
For natural language, this is a reasonable inductive bias: nearby tokens in the sequence are often semantically or syntactically related, and sequence recency provides a useful approximation to contextual relevance.

This assumption becomes problematic for raster-scanned image tokens. 
Within a row, adjacent sequence positions correspond to horizontally adjacent image tokens. 
At a row transition, however, the last token of row $r$ and the first token of row $r+1$ are adjacent in the flattened sequence but not adjacent in the 2D grid. 
Thus, a 1D decay mechanism treats an artificial sequence neighbor as a highly local visual neighbor. 
This creates a \emph{false-locality} bias: information propagation is guided by sequence distance even when sequence distance disagrees with spatial distance.

This issue also appears in advanced decay-based linear attention. 
For example, HGRN2~\cite{qin2024hgrn2} associates the input gate with the key and uses the complementary gate as the retention factor:
\begin{equation}
\begin{gathered}
\mathbf{s}_0=\mathbf{0},\qquad
\mathbf{s}_t=\mathrm{diag}(\lambda_t)\mathbf{s}_{t-1}
+(1-\lambda_t)\mathbf{v}_t^\top,\qquad
\mathbf{o}_t^\top=\mathbf{q}_t^\top\mathbf{s}_t,\\
\mathbf{q}_t,\mathbf{v}_t,\mathbf{o}_t,\lambda_t\in\mathbb{R}^{d},\qquad
\mathbf{s}_t\in\mathbb{R}^{d\times d},\qquad
0\leq\lambda_t\leq1,\quad t=1,\ldots,N.
\label{equ:hgrn2}
\end{gathered}
\end{equation}
Although such mechanisms are effective for 1D sequence modeling, our experiments in~Table~\ref{tab:abl_arch_eval,tab:abl_decay_eval} show that directly applying them to image generation remains suboptimal. 
The missing component is not merely stronger recurrence, but a decay rule that is aware of the original 2D token geometry.

\begin{algorithm}[t]
\caption{Linear Attention with Spatial-Aware Decay.}
\label{alg:decay}
\begin{algorithmic}[1]
\STATE \textbf{Input:} sequence $\mathbf{X}\in\mathbb{R}^{B\times N\times D}$, token-grid width $w$
\STATE \textbf{Output:} sequence $\mathbf{O}\in\mathbb{R}^{B\times N\times D}$
\STATE $\mathbf{s}_0\gets\mathbf{0}$
\FOR{$t=1$ to $N$}
    \STATE $\mathbf{q}_t,\mathbf{v}_t,\lambda_t\gets\mathrm{Linear}(\mathbf{x}_t)$
    \STATE $\mathbf{q}_t\gets\mathrm{SiLU}(\mathbf{q}_t)$,\quad $\lambda_t\gets\mathrm{Sigmoid}(\lambda_t)$
    \STATE $\mathbf{k}_t\gets1-\lambda_t$
    \STATE $m_t\gets\mathbb{I}(t\bmod w\neq0)$
    \STATE $\log\lambda_t^{\mathrm{spatial}}\gets m_t\log\lambda_t$
    \STATE $\mathbf{s}_t\gets\mathrm{diag}(\lambda_t^{\mathrm{spatial}})\mathbf{s}_{t-1}+\mathbf{k}_t\mathbf{v}_t^\top$
    \STATE $\mathbf{o}_t^\top\gets\mathbf{q}_t^\top\mathbf{s}_t$
    \STATE $\mathbf{o}_t\gets\mathrm{Linear}(\mathrm{Norm}(\mathbf{o}_t))$
\ENDFOR
\STATE \textbf{return} $\mathbf{O}=[\mathbf{o}_1,\mathbf{o}_2,\ldots,\mathbf{o}_N]$
\end{algorithmic}
\end{algorithm}

\subsection{Linear Attention with Spatial-Aware Decay}
\label{sec:sec3.4}

We propose Linear Attention with Spatial-Aware Decay (LASAD) to align recurrent linear attention with the 2D structure of image tokens. 
The central idea is simple: the decay applied between consecutive sequence positions should depend not only on their 1D indices, but also on whether the transition corresponds to a valid local transition in the original token grid. 
In particular, row-boundary transitions are artificial neighbors introduced by raster scanning and should not be processed by the same decay rule as within-row transitions.

Given the recurrent state formulation, LASAD updates the state as
\begin{equation}
\begin{gathered}
\mathbf{s}_0=\mathbf{0},\qquad
\mathbf{s}_t=\mathrm{diag}(\lambda_t^{\text{spatial}})\mathbf{s}_{t-1}
+\mathbf{k}_t\mathbf{v}_t^\top,\qquad
\mathbf{o}_t^\top=\mathbf{q}_t^\top\mathbf{s}_t,\\
\mathbf{q}_t,\mathbf{k}_t,\mathbf{v}_t,\mathbf{o}_t,\lambda_t^{\text{spatial}}\in\mathbb{R}^{d},\qquad
\mathbf{s}_t\in\mathbb{R}^{d\times d},\qquad
0\leq\lambda_t^{\text{spatial}}\leq1.
\label{equ:lasad}
\end{gathered}
\end{equation}
We first compute a content-dependent base retention factor from the input gate:
\begin{equation}
\lambda_t=1-\mathbf{k}_t.
\end{equation}
Then we introduce a row-boundary mask
\begin{equation}
m_t=\mathbb{I}(t\bmod w\neq0),
\end{equation}
where $w$ is the width of the token grid. 
The spatial-aware decay is defined in log space:
\begin{equation}
\log\lambda_t^{\text{spatial}}=m_t\log\lambda_t,\qquad t=1,\ldots,N.
\label{equ:sad}
\end{equation}
For normal within-row transitions, $m_t=1$ and LASAD uses the learned decay $\lambda_t$. 
At row-boundary positions, $m_t=0$, hence $\log\lambda_t^{\text{spatial}}=0$ and equivalently $\lambda_t^{\text{spatial}}=1$. 
This neutralizes the ordinary 1D decay at artificial row transitions, so the recurrent update is not biased by false sequence locality. 
Importantly, the mechanism does not introduce a hand-tuned decay strength; it only uses the known raster-scan geometry to decide where the learned decay should be spatially adjusted.

\subsection{LASADGen Architecture}
\label{sec:lasadgen_arch}

LASADGen follows the autoregressive generation pipeline of LlamaGen~\cite{llamagen}, but replaces the softmax-attention modules with LASAD layers. 
An input image is first tokenized into a discrete 2D token grid and flattened into a raster-scanned sequence. 
During training, the class condition and previous visual tokens are fed into a stack of $L$ LASAD blocks to predict the next visual token. 
During inference, tokens are generated sequentially, and each LASAD block maintains its recurrent state $\mathbf{s}_t$ instead of storing an ever-growing KV cache. 
This preserves the standard AR next-token prediction interface while changing the internal sequence mixer from quadratic softmax attention to spatial-aware recurrent linear attention.

Each LASAD block consists of a projection layer that produces the query, value, and retention variables; the spatial-aware decay update in~Equ.\ref{equ:lasad,equ:sad}; a normalization layer; and an output projection. 
The block is therefore a drop-in replacement for the attention module in an AR generator. 
The only additional information required by LASAD is the token-grid width $w$, which is already determined by the tokenizer and image resolution.



\begin{table}[t]
\centering
\small
\setlength{\tabcolsep}{0.16cm}
\caption{\textbf{Details of LASAD models.} We follow the configurations of LlamaGen~\cite{llamagen} and get different variants.}
\begin{tabular}{lcccc}
\toprule
\textbf{Model}     & \textbf{Layers} $L$ & \textbf{Hidden size} $d$ & \textbf{Heads} & \textbf{\#Para.} \\ \midrule
LASAD-B            & 12         & 768             & 12    & 111M   \\
LASAD-L            & 24         & 1024            & 16    & 346M  \\
LASAD-XL           & 36         & 1280            & 20    & 774M   \\
LASAD-XXL          & 48         & 1536            & 24    & 1.4B   \\
\bottomrule
\end{tabular}
\label{tab:config}
\end{table}

\begin{table}[h]
\centering
\small
\setlength{\tabcolsep}{4.pt}
\renewcommand\arraystretch{1.0}
\setlength{\abovecaptionskip}{0.1cm}
\caption{\textbf{Quantitative comparisons} of different architectures for autoregressive image generation. \enquote{LinearAttn} denotes the original linear attention augmented with RoPE~\cite{su2024roformer}.
\enquote{Hybrid-Linear} refers to the hybrid architecture that integrates one softmax attention layer within the LinearAttn framework.
\enquote{SoftmaxAttn} represents the conventional softmax-based attention mechanism.
TNL and GLA represent two decay-based methods respectively.}
\begin{tabular}{clcccc}
\toprule
\textbf{Size} & \textbf{Model} & \textbf{FID}$\downarrow$ & \textbf{IS}$\uparrow$ & \textbf{Precision}$\uparrow$ & \textbf{Recall}$\uparrow$ \\
\midrule
\multirow{6}{*}{B}
& LinearAttn~\cite{katharopoulos_transformers_are_rnn_ICML_2020} & 50.86 & 19.27 & 0.45 & 0.40 \\
& SoftmaxAttn & 5.46 & 193.61 & \textbf{0.83} & 0.45 \\
& Hybrid-Linear & 6.36 & 163.65 & 0.81 & 0.46 \\
& TNL~\cite{qin_tnl_various} & 6.75 & 161.92 & 0.80 & 0.46 \\
& GLA~\cite{yang_GLA_2023} & 6.45 & 163.21 & 0.80 & 0.47 \\
\cmidrule(lr){2-6}
& LASADGen & \textbf{4.86} & \textbf{194.91} & 0.82 & \textbf{0.48} \\
\midrule
\multirow{6}{*}{L}

& LinearAttn~\cite{katharopoulos_transformers_are_rnn_ICML_2020} & 37.50 & 28.99 & 0.53 & 0.43 \\
& SoftmaxAttn                & 3.80 & 248.28 & \textbf{0.83} & 0.51 \\
& Hybrid-Linear               & 3.64 & 224.67 & 0.81 & 0.52 \\
& TNL~\cite{qin_tnl_various} & 3.61 & 228.51 & 0.81 & 0.51 \\
& GLA~\cite{yang_GLA_2023}   & 3.58 & 234.21 & 0.81 & 0.51 \\
\cmidrule(lr){2-6}
& LASADGen                   & \textbf{2.90} & \textbf{255.61} & 0.82 & \textbf{0.55} \\
\bottomrule
\end{tabular}
\label{tab:abl_arch_eval}
\end{table}

\section{Experiments}
\subsection{Implementation Details}

\noindent\textbf{Benchmark.}
We develop LASADGen by extending the LlamaGen architecture~\cite{llamagen}, since it is a simple and scalable framework for autoregressive image generation. 
For comprehensive evaluation, we conduct experiments on class-conditional image generation using the ImageNet1K~\cite{deng2009imagenet}, which serves as a standard benchmark in the field.

\noindent\textbf{Evaluation Metrics.} 
We evaluate our method using the Fréchet Inception Distance (FID)~\cite{heusel_fid_2017} as the primary metric, complemented by the Inception Score (IS)~\cite{salimans_IS_2016} and precision-recall metrics~\cite{kynkaanniemi2019improved}. 
For fair comparison, we directly cite baseline results from their respective original papers. All quantitative evaluations are conducted using the official TensorFlow implementation from ADM~\cite{dhariwal_adm_nips_2021diffusion} to ensure consistent measurement conditions.

\noindent\textbf{Training Details.}
We conduct our experiments trained on ImageNet1K~\cite{deng2009imagenet} at $256\times 256$ resolution. 
The training is performed on 8 NVIDIA H800 GPUs, with each image being discretized into 256 tokens. 
For optimization, we employ AdamW with momentum parameters $(\beta_1, \beta_2) = (0.9, 0.95)$ and a weight decay coefficient of 0.05. 
The learning rate is set to $1\times 10^{-4}$ for a batch size of 256, and the models are trained for 300 epochs. 
To enable classifier-free guidance~\cite{ho_cfg_2022} during inference, we applied a dropout rate of 0.1 to the class embedding layers. 
The image tokenizer incorporates a downsampling factor of 16 and leverages pre-trained weights from LlamaGen~\cite{llamagen}.
We use the \texttt{Flash Linear Attention} library~\cite{yang2024fla} to get an efficient linear attention implementation.

\subsection{Model Configurations}
The proposed LASADGen model consists of $L$ LASAD blocks, each with the hidden dimension size $d$. 
Consistent with LlamaGen~\cite{llamagen}, we adopt standard transformer configs that jointly scale $L$, $d$ and the number of attention heads. 
Specifically, we establish four configs: LASAD-B, LASAD-L, LASAD-XL and LASAD-XXL. 
These configurations span a broad range of model sizes and parameter distributions, ranging from 111M to 1.4B parameters, enabling us to evaluate the impact of scaling. 
Detailed information about these configurations is provided in~Table~\ref{tab:config}.


\begin{table*}[t]
    \centering
    \small
    \setlength{\tabcolsep}{6pt}
    \renewcommand\arraystretch{1.0}
    \setlength{\abovecaptionskip}{0.1cm}
    \caption{\textbf{Class-conditional generation on $256 \times 256$ ImageNet.} 
    $*$ specifies the generated images are $384 \times 384$ and are resized to 256×256 for evaluation. 
    \enquote{-re} denotes rejection sampling is used.
    $^{\dagger}$ indicates the performance after only 50 epochs of training.
    The evaluation protocol and implementation are the same with ADM.}
    \begin{tabular}{lccccc}
    \toprule
    \textbf{Model} & \textbf{\#Para.} & \textbf{FID}$\downarrow$ & \textbf{IS}$\uparrow$ & \textbf{Precision}$\uparrow$ & \textbf{Recall}$\uparrow$  \\
    \midrule
    VQGAN~\cite{vqgan} & 227M & 18.65 & 80.4         & 0.78 & 0.26    \\
     VQGAN~\cite{vqgan}    & 1.4B   & 15.78 & 74.3   & $-$  & $-$     \\
     VQGAN-re~\cite{vqgan}  & 1.4B  & 5.20  & 280.3  & $-$  & $-$     \\
     ViT-VQGAN~\cite{vit-vqgan} & 1.7B & 4.17  & 175.1  & $-$  & $-$        \\
     ViT-VQGAN-re~\cite{vit-vqgan}& 1.7B  & 3.04  & 227.4  & $-$  & $-$     \\
     RQTran.~\cite{rqvae}       & 3.8B  & 7.55  & 134.0  & $-$  & $-$     \\
     RQTran.-re~\cite{rqvae}    & 3.8B & 3.80  & 323.7  & $-$  & $-$    \\
    \midrule
     LlamaGen-B$^{*}$~\cite{llamagen} & 111M & 6.43 & 157.17 & 0.81 & 0.46 \\
     LlamaGen-L$^{*}$~\cite{llamagen} & 343M & 3.07 & 256.06 & 0.83 & 0.52 \\
     LlamaGen-XL$^{*}$~\cite{llamagen} & 775M & 2.62 & 244.08 & 0.80 & 0.57 \\
     LlamaGen-XXL$^{*}$~\cite{llamagen} & 1.4B & 2.34 & 253.90 & 0.80 & 0.59 \\
     LlamaGen-B~\cite{llamagen} & 111M & 5.46 & 193.61 & 0.83 & 0.45 \\
     LlamaGen-L~\cite{llamagen} & 343M & 3.80 & 248.28 & 0.83 & 0.51 \\
     LlamaGen-XL$^{\dagger}$~\cite{llamagen} & 775M & 3.39 & 227.08 & 0.81 & 0.54 \\
     LlamaGen-XXL$^{\dagger}$~\cite{llamagen} & 1.4B & 3.09 & 253.61 & 0.83 & 0.53 \\
    \midrule
     LASADGen-B  & 112M  & 4.86 & 194.91 & 0.82 & 0.48  \\
     LASADGen-L  & 346M  & 2.90 & 255.61 & 0.82 & 0.55 \\
     LASADGen-XL & 774M  & 2.58 & 282.73 & 0.82 & 0.56 \\
     LASADGen-XXL & 1.4B & 2.36 & 283.2 & 0.82 & 0.57  \\
     LASADGen-XL$^{\dagger}$ & 774M  & 3.07 & 269.35 & 0.83 & 0.54 \\
     LASADGen-XXL$^{\dagger}$ & 1.4B & 2.81 & 270.42 & 0.83 & 0.55  \\
    \bottomrule
    \end{tabular}
    \label{tab:main_table}
\end{table*}

\subsection{Ablation of Ordinary Linear Attention}
We first conduct experiments to answer the question posed in~Sec.\ref{sec:intro}: \enquote{\textit{Can linear attention be effectively applied to autoregressive image generation models without compromising performance?}}.

In Table \ref{tab:abl_arch_eval}, we present the performance of various linear attention variants on autoregressive image generation tasks. 
We first examine two baseline experiments: \enquote{LinearAttn} and \enquote{Hybrid-Linear}. 
The former implements basic linear attention by directly substituting kernel functions for the softmax operation~\cite{katharopoulos_transformers_are_rnn_ICML_2020}, while the latter employs a hybrid architecture \cite{li2025minimax} that intersperses only one standard attention layer among linear attention layers. 
Our results indicate that basic linear attention fails entirely to adapt to this task.
The hybrid attention model demonstrates substantial performance improvements over basic linear attention by effectively combining softmax attention's superior long-range dependency modeling capabilities with linear attention's computational efficiency.
However, despite these advantages, hybrid model still underperforms compared to our proposed LASAD approach.
We further investigate two decay-based linear attention mechanisms: TNL \cite{qin_tnl_various} and GLA \cite{yang_GLA_2023}. Our analysis reveals that these models demonstrate notable performance deficiencies when compared to our proposed LASAD approach. These experimental results indicate that advanced linear attention architectures incorporating decay mechanisms still fail to achieve optimal efficacy in autoregressive image generation tasks, although they have been validated on language models.

\subsection{Performance Comparison}
\label{sub_sec_performance_comparison}
\noindent\textbf{Quantitative Comparison.}
We compare the performance of our proposed LASADGen with prior image generation methods, as presented in~Table~\ref{tab:main_table}. 
To demonstrate the generality of our model, we evaluate results across four different behavior. 
Our method consistently achieves significantly superior image generation quality compared to state-of-the-art autoregressive methods, highlighting its robustness and effectiveness. 
Notably, LASADGen outperforms LlamaGen across all model sizes tested. 
Moreover, our model trained at $256\times 256$ resolution surpasses the performance of LlamaGen at $384\times 384$ resolution. 

\begin{figure}[h]
  \centering
  \setlength{\abovecaptionskip}{0.2cm}
    \includegraphics[width=0.5\linewidth]{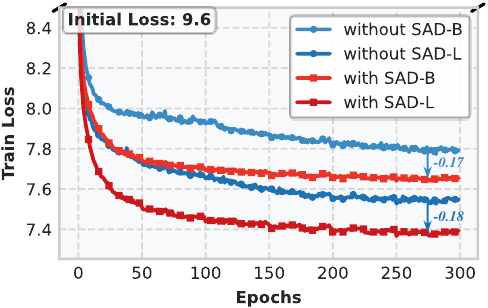}
    \caption{\textbf{Training loss comparison between the model with or without the proposed Spatial-Aware Decay}. We visualize the training process of the two groups of models in~Table\ref{tab:abl_decay_eval}. The blue curve is used to represent the model without using SAD, and the red curve is used to represent the model using SAD.}
    \label{fig:loss}
\end{figure}

Furthermore, for a more complete comparison with LlamaGen, which does not report results at 300 epochs for their XL and XXL models (trained at $256\times 256$ resolution), we present additional results for our LASADGen at 50 epochs using the XL and XXL models. 
As indicated by the results marked with $^{\dagger}$ in~Table~\ref{tab:main_table}, LASADGen demonstrates superior performance even under shorter training epochs, further validating its effectiveness.
This can also be demonstrated in~Figure~\ref{fig:loss}, which shows that our LASADGen exhibits strong convergence ability in the first 50 epochs.


\begin{figure*}[t]
    \centering
    \setlength{\abovecaptionskip}{0.2cm}
    \includegraphics[width=0.9\textwidth]{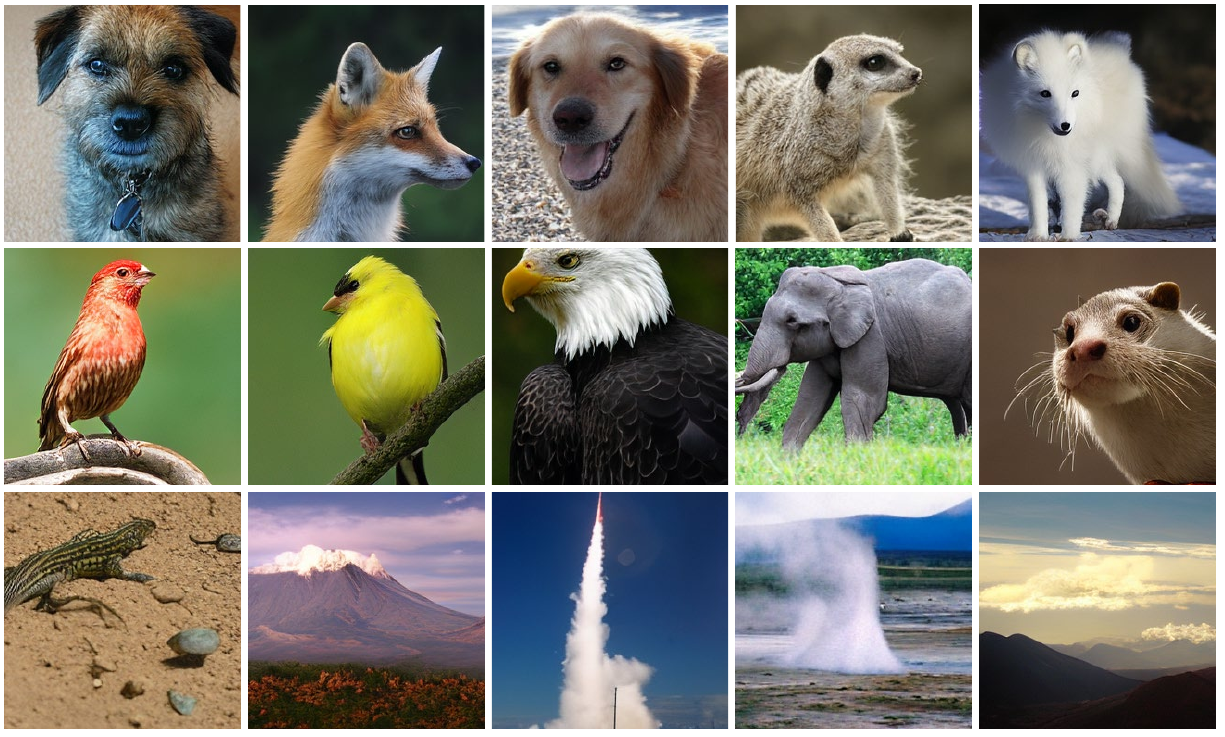}
    \caption{\textbf{Autoregressive Image Generation with LASADGen.} We show the samples from our class-conditional LASADGen-XL model trained on ImageNet at $256\times 256$ resolution.}
    \label{fig:vis}
\end{figure*}

\noindent\textbf{Visualization.} 
As shown in~Table~\ref{fig:vis}, we visualize some sample images randomly generated by our LASADGen. 
We can observe that LASADGen is capable of generating high-quality samples with both high fidelity and diversity.
The generated images exhibit remarkable structural coherence and fine-grained details, particularly in complex textures and object boundaries.
These visualizations substantiate the effectiveness of our proposed architectural improvements by the LASADGen.

\noindent\textbf{Efficiency Analysis.}
Table~\ref{tab:inference_time} reports the inference time comparison between our proposed LASADGen and the autoregressive baseline LlamaGen across different model scales. 
As shown in the table, LASADGen consistently achieves faster inference speeds at all scales. 
Overall, LASADGen provides approximately $1.6\times$--$1.7\times$ speedup compared with LlamaGen. 
This efficiency gain mainly stems from the linear computational complexity of our architecture, which reduces the cost of token interactions during generation.
For the training time, the original LlamaGen requires 9.62s, whereas our model takes 8.31s, demonstrating efficiency under the same training setup.
Furthermore, our linear-complexity architecture eliminates the need to store a key-value (KV) cache during autoregressive inference, enabling long-sequence modeling without increasing memory consumption. 
Consistently, LASADGen requires only 819.07 MB of GPU memory during inference, which is significantly lower than the 1439.63 MB consumed by LlamaGen.

\begin{table}[h]
    \centering
    \small
    \caption{Inference time (seconds) comparison across model scales.}
    \label{tab:inference_time}
    \begin{tabular}{lcccc}
    \toprule
    \textbf{Model} & \textbf{B} & \textbf{L} & \textbf{XL} & \textbf{XXL} \\
    \midrule
    LlamaGen (vllm) & 0.610 & 1.150 & 1.860 & 2.760 \\
    LASADGen        & \textbf{0.363} & \textbf{0.691} & \textbf{1.116} & \textbf{1.659} \\
    \bottomrule
    \end{tabular}
\end{table}

\begin{figure*}[h]
    \centering
    \begin{minipage}[b]{0.48\textwidth}
        \centering
        \includegraphics[width=\textwidth]{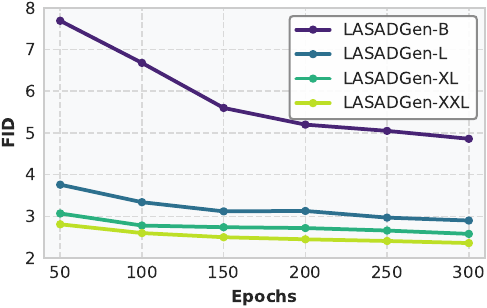}
        \label{fig:scaling-metrics}
    \end{minipage}
    \hfill
    \begin{minipage}[b]{0.48\textwidth}
        \centering
        \includegraphics[width=\textwidth]{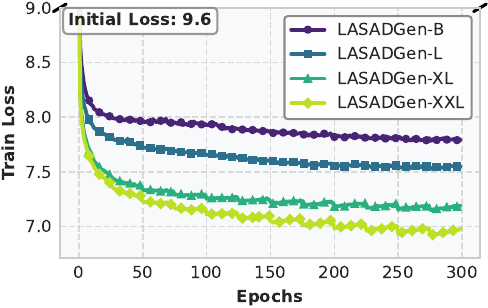}
        \label{fig:scaling-visual}
    \end{minipage}
    \caption{\textbf{Scaling properties of LASADGen.} \textbf{Left:} FID scores and \textbf{Right:} Training loss vs. model size.}
    \label{fig:scaling}
\end{figure*}


\noindent\textbf{Scaling Model Scales.}
We investigate the scaling behavior of LASADGen. As shown in~Figure~\ref{fig:scaling}, we show the training loss and FID score based on models of different sizes on different training iterations. 
It can be observed that LASADGen demonstrates good scalability, with lower training loss and better FID scores as the model size increases. 
This is because we did not modify the structure of the AR model itself, preserving the integrity of the AR framework, and therefore inheriting the scalability of AR methods. 

\noindent\textbf{Long Sequence Scalability.}
To demonstrate long sequence scalability, we evaluate LASADGen-L across different resolutions and sequence lengths. We test at 384$\times$384 resolution (576 tokens) and with longer sequences up to 1024 tokens using an 8$\times$ downsampling tokenizer. As shown in~Table~\ref{tab:scalability}, our method consistently outperforms LlamaGen-L across all settings while maintaining linear computational complexity.

\begin{table}[!tbp]
    \centering
    \small
    \setlength{\tabcolsep}{2pt}
    \renewcommand\arraystretch{1.0}
    \setlength{\abovecaptionskip}{0.1cm}
    \caption{\textbf{Long sequence scalability} with different resolutions and sequence lengths.}
    \begin{tabular}{lccccccc}
    \toprule
    \textbf{Model} & \textbf{Size} & \textbf{Tokens} & \textbf{FID}$\downarrow$ & \textbf{IS}$\uparrow$ & \textbf{Precision}$\uparrow$ & \textbf{Recall}$\uparrow$ \\
    \midrule
    LlamaGen-L & 384$^2$ & 576 & 3.07 & 256.06 & 0.83 & 0.52 \\
    LASADGen-L & 384$^2$ & 576 & \textbf{2.78} & \textbf{265.78} & \textbf{0.83} & \textbf{0.55} \\
    \midrule
    LlamaGen-L & 256$^2$ & 1024 & 2.95 & 262.31 & 0.83 & 0.53 \\
    LASADGen-L & 256$^2$ & 1024 & \textbf{2.71} & \textbf{271.45} & \textbf{0.83} & \textbf{0.56} \\
    \bottomrule
    \end{tabular}
    
    \label{tab:scalability}
\end{table}

\subsection{Ablations of the Spatial-Aware Decay}
To validate the effectiveness of our proposed Spatial-Aware Decay (SAD) mechanism, we conduct ablation studies on both base (B) and large (L) model variants. As shown in~Figure~\ref{fig:loss}, incorporating SAD consistently leads to improved training dynamics across model scales. 
Both LASADGen-B and LASADGen-L variants demonstrate more efficient optimization, achieving lower training loss values throughout the training process. Particularly noteworthy is the significant gap between models with and without SAD after convergence. 
This improvement in optimization efficiency directly translates to enhanced generation quality, as evidenced by the quantitative metrics~Table~\ref{tab:abl_decay_eval}.
We observe substantial improvements in FID scores for both model sizes, with similar gains in Inception Score. 
The precision and recall metrics also show consistent improvements, indicating that SAD enhances both the fidelity and diversity of generated samples. 
Furthermore, we apply our proposed SAD to GLA-B~\cite{yang_GLA_2023} and conduct experiments. 
The experimental results show that SAD can also produce performance gains for GLA.
These results confirm that our proposed Spatial-Aware Decay mechanism effectively addresses the limitations of the baseline linear attention model, leading to more coherent and visually appealing image generation.


\begin{table}[!tbp]
    \centering
    \small
    \setlength{\tabcolsep}{3.2pt}
    \renewcommand\arraystretch{1.0}
    \setlength{\abovecaptionskip}{0.1cm}
    \caption{\textbf{Quantitative comparisons} of the model with or without the SAD.}
    \begin{tabular}{clccccc}
    \toprule
    \textbf{Size} & \textbf{Model} & \textbf{SAD} & \textbf{FID}$\downarrow$ & \textbf{IS}$\uparrow$ & \textbf{Precision}$\uparrow$ & \textbf{Recall}$\uparrow$ \\
    \midrule
    \multirow{4}{*}{B}
    & GLA~\cite{yang_GLA_2023} & \ding{55} & 6.45 & 163.21 & 0.80 & 0.47 \\
    & GLA~\cite{yang_GLA_2023} & \ding{51} & 5.84 & 173.00 & 0.80 & 0.48 \\
    \cmidrule(lr){2-7}
    & LASADGen & \ding{55} & 7.42 & 151.02 & 0.78 & 0.47 \\
    & LASADGen & \ding{51} & \textbf{4.86} & \textbf{194.91} & \textbf{0.82} & \textbf{0.48} \\
    \midrule
    \multirow{2}{*}{L}
    & LASADGen & \ding{55} & 3.73 & 231.88 & 0.81 & 0.51 \\
    & LASADGen & \ding{51} & \textbf{2.90} & \textbf{255.61} & \textbf{0.82} & \textbf{0.55} \\
    \bottomrule
    \end{tabular}
    \label{tab:abl_decay_eval}
\end{table}


\section{Conclusion}
\label{sec:conclusion}

In this paper, we introduced LASADGen, a novel autoregressive image generation framework powered by Linear Attention with Spatial-Aware Decay (LASAD). Our method effectively preserves 2D spatial relationships within flattened sequences, overcoming critical limitations of existing linear attention mechanisms originally designed for 1D sequential data. 
Experimental results on ImageNet demonstrate that LASADGen achieves state-of-the-art image generation quality while significantly reducing computational complexity. 
This work opens promising directions for efficient autoregressive modeling of spatial contexts in visual data and highlights the potential of spatial-aware attention.
Future research directions include adapting LASAD for variable-resolution generation and expanding its application to larger-scale tasks, such as text-to-image generation or video generation.


\section*{Acknowledgements}
{This research was supported in part by National Natural Science Foundation of China (62525115, U2570208).}

\bibliographystyle{scis}
\bibliography{main}

@String(ICLR  = {International Conference on Learning Representations (ICLR)})

@String(AAAI  = {AAAI Conference on Artificial Intelligence (AAAI)})

@String{SIGGRAPH = {ACM SIGGRAPH}}

@String(CVPR  = {Conference on Computer Vision and Pattern Recognition (CVPR)})

@String(ICCV  = {International Conference on Computer Vision (ICCV)})

@String(NIPS  = {Conference on Neural Information Processing Systems (NeurIPS)})

@String(TIP   = {IEEE Transactions on Image Processing (IEEE TIP)})

@String(TCSVT = {IEEE Transactions on Circuits and Systems for Video Technology (IEEE TCSVT)})

@String(TMM   = {IEEE Transactions on Multimedia (IEEE TMM)})

@String(ACMMM = {ACM Multimedia Conference (ACM MM)})

@String(ACL   = {Association for Computational Linguistics (ACL)})

@String(EMNLP = {Conference on Empirical Methods in Natural Language Processing (EMNLP)})

@String(ICML  = {International Conference on Machine Learning (ICML)})

@String(JMLR  = {Journal of Machine Learning Research (JMLR)})

@article{biggan,
  title={Large scale GAN training for high fidelity natural image synthesis},
  author={Brock, Andrew and Donahue, Jeff and Simonyan, Karen},
  journal={arXiv preprint arXiv:1809.11096},
  year={2018}
}

@inproceedings{gigagan,
  title={Scaling up gans for text-to-image synthesis},
  author={Kang, Minguk and Zhu, Jun-Yan and Zhang, Richard and Park, Jaesik and Shechtman, Eli and Paris, Sylvain and Park, Taesung},
  booktitle=CVPR,
  pages={10124--10134},
  year={2023}
}

@inproceedings{stylegan-xl,
  title={Stylegan-xl: Scaling stylegan to large diverse datasets},
  author={Sauer, Axel and Schwarz, Katja and Geiger, Andreas},
  booktitle=SIGGRAPH,
  pages={1--10},
  year={2022}
}

@article{cdm,
  title={Cascaded diffusion models for high fidelity image generation},
  author={Ho, Jonathan and Saharia, Chitwan and Chan, William and Fleet, David J and Norouzi, Mohammad and Salimans, Tim},
  journal=JMLR,
  volume={23},
  number={1},
  pages={2249--2281},
  year={2022},
}

@inproceedings{ldm,
  title={High-resolution image synthesis with latent diffusion models},
  author={Rombach, Robin and Blattmann, Andreas and Lorenz, Dominik and Esser, Patrick and Ommer, Bj{\"o}rn},
  booktitle=CVPR,
  pages={10684--10695},
  year={2022}
}

@inproceedings{dit,
  title={Scalable diffusion models with transformers},
  author={Peebles, William and Xie, Saining},
  booktitle=CVPR,
  pages={4195--4205},
  year={2023}
}

@inproceedings{vqgan,
  author       = {Patrick Esser and
                  Robin Rombach and
                  Bj{\"{o}}rn Ommer},
  title        = {Taming Transformers for High-Resolution Image Synthesis},
  booktitle    = CVPR,
  pages        = {12873--12883},
  year         = {2021},
}

@inproceedings{vit-vqgan,
  author       = {Jiahui Yu and
                  Xin Li and
                  Jing Yu Koh and
                  Han Zhang and
                  Ruoming Pang and
                  James Qin and
                  Alexander Ku and
                  Yuanzhong Xu and
                  Jason Baldridge and
                  Yonghui Wu},
  title        = {Vector-quantized Image Modeling with Improved {VQGAN}},
  booktitle    = ICLR,
  year         = {2022},
}

@inproceedings{rqvae,
  author       = {Doyup Lee and
                  Chiheon Kim and
                  Saehoon Kim and
                  Minsu Cho and
                  Wook{-}Shin Han},
  title        = {Autoregressive Image Generation using Residual Quantization},
  booktitle    = CVPR,
  pages        = {11513--11522},
  year         = {2022},
}

@article{llamagen,
  title={Autoregressive Model Beats Diffusion: Llama for Scalable Image Generation},
  author={Sun, Peize and Jiang, Yi and Chen, Shoufa and Zhang, Shilong and Peng, Bingyue and Luo, Ping and Yuan, Zehuan},
  journal={arXiv preprint arXiv:2406.06525},
  year={2024}
}

@article{var,
  title={Visual autoregressive modeling: Scalable image generation via next-scale prediction},
  author={Tian, Keyu and Jiang, Yi and Yuan, Zehuan and Peng, Bingyue and Wang, Liwei},
  journal={arXiv preprint arXiv:2404.02905},
  year={2024}
}

@inproceedings{katharopoulos_transformers_are_rnn_ICML_2020,
  title={Transformers are rnns: Fast autoregressive transformers with linear attention},
  author={Katharopoulos, Angelos and Vyas, Apoorv and Pappas, Nikolaos and Fleuret, Fran{\c{c}}ois},
  booktitle=ICML,
  pages={5156--5165},
  year={2020}
}

@inproceedings{peng2020random,
  title={Random Feature Attention},
  author={Peng, Hao and Pappas, Nikolaos and Yogatama, Dani and Schwartz, Roy and Smith, Noah and Kong, Lingpeng},
  booktitle=ICLR,
  year={2020}
}

@inproceedings{choromanski2020rethinking,
  title={Rethinking Attention with Performers},
  author={Choromanski, Krzysztof Marcin and Likhosherstov, Valerii and Dohan, David and Song, Xingyou and Gane, Andreea and Sarlos, Tamas and Hawkins, Peter and Davis, Jared Quincy and Mohiuddin, Afroz and Kaiser, Lukasz and others},
  booktitle=ICLR,
  year={2020}
}

@inproceedings{qin_iclr_cosformer,
  title={cosFormer: Rethinking Softmax In Attention},
  author={Qin, Zhen and Sun, Weixuan and Deng, Hui and Li, Dongxu and Wei, Yunshen and Lv, Baohong and Yan, Junjie and Kong, Lingpeng and Zhong, Yiran},
  booktitle=ICLR,
  year={2022}
}

@inproceedings{qin_tnl_various,
  title={Various Lengths, Constant Speed: Efficient Language Modeling with Lightning Attention},
  author={Qin, Zhen and Sun, Weigao and Li, Dong and Shen, Xuyang and Sun, Weixuan and Zhong, Yiran},
  booktitle=ICML,
  year={2024}
}

@inproceedings{chang2022maskgit,
  title={Maskgit: Masked generative image transformer},
  author={Chang, Huiwen and Zhang, Han and Jiang, Lu and Liu, Ce and Freeman, William T},
  booktitle=CVPR,
  pages={11315--11325},
  year={2022}
}

@inproceedings{li_Mage_cvpr_2023,
  title={Mage: Masked generative encoder to unify representation learning and image synthesis},
  author={Li, Tianhong and Chang, Huiwen and Mishra, Shlok and Zhang, Han and Katabi, Dina and Krishnan, Dilip},
  booktitle=CVPR,
  pages={2142--2152},
  year={2023}
}

@article{li_MAR_nips_2025,
  title={Autoregressive image generation without vector quantization},
  author={Li, Tianhong and Tian, Yonglong and Li, He and Deng, Mingyang and He, Kaiming},
  journal=NIPS,
  volume={37},
  pages={56424--56445},
  year={2025}
}

@inproceedings{yu2023magvit,
  title={Magvit: Masked generative video transformer},
  author={Yu, Lijun and Cheng, Yong and Sohn, Kihyuk and Lezama, Jos{\'e} and Zhang, Han and Chang, Huiwen and Hauptmann, Alexander G and Yang, Ming-Hsuan and Hao, Yuan and Essa, Irfan and others},
  booktitle=CVPR,
  pages={10459--10469},
  year={2023}
}

@inproceedings{devlin2019bert,
  title={Bert: Pre-training of deep bidirectional transformers for language understanding},
  author={Devlin, Jacob and Chang, Ming-Wei and Lee, Kenton and Toutanova, Kristina},
  booktitle=ACL,
  pages={4171--4186},
  year={2019}
}

@inproceedings{esser2021taming,
  title={Taming transformers for high-resolution image synthesis},
  author={Esser, Patrick and Rombach, Robin and Ommer, Bjorn},
  booktitle=CVPR,
  pages={12873--12883},
  year={2021}
}

@article{yu2021vector,
  title={Vector-quantized image modeling with improved vqgan},
  author={Yu, Jiahui and Li, Xin and Koh, Jing Yu and Zhang, Han and Pang, Ruoming and Qin, James and Ku, Alexander and Xu, Yuanzhong and Baldridge, Jason and Wu, Yonghui},
  journal={arXiv preprint arXiv:2110.04627},
  year={2021}
}

@article{radford2018improving,
  title={Improving language understanding by generative pre-training},
  author={Radford, Alec and Narasimhan, Karthik and Salimans, Tim and Sutskever, Ilya and others},
  journal={OpenAI technical report},
  year={2018},
  publisher={San Francisco, CA, USA}
}

@article{kingma2013auto,
  title={Auto-Encoding Variational Bayes},
  author={Kingma, Diederik P and Welling, Max},
  journal={arXiv preprint arXiv:1312.6114},
  year={2013}
}

@article{van2017neural,
  title={Neural discrete representation learning},
  author={Van Den Oord, Aaron and Vinyals, Oriol and others},
  journal=NIPS,
  volume={30},
  year={2017}
}

@article{touvron2023llama,
  title={Llama: Open and efficient foundation language models},
  author={Touvron, Hugo and Lavril, Thibaut and Izacard, Gautier and Martinet, Xavier and Lachaux, Marie-Anne and Lacroix, Timoth{\'e}e and Rozi{\`e}re, Baptiste and Goyal, Naman and Hambro, Eric and Azhar, Faisal and others},
  journal={arXiv preprint arXiv:2302.13971},
  year={2023}
}

@article{heusel_fid_2017,
  title={Gans trained by a two time-scale update rule converge to a local nash equilibrium},
  author={Heusel, Martin and Ramsauer, Hubert and Unterthiner, Thomas and Nessler, Bernhard and Hochreiter, Sepp},
  journal=NIPS,
  volume={30},
  year={2017}
}

@article{salimans_IS_2016,
  title={Improved techniques for training gans},
  author={Salimans, Tim and Goodfellow, Ian and Zaremba, Wojciech and Cheung, Vicki and Radford, Alec and Chen, Xi},
  journal=NIPS,
  volume={29},
  year={2016}
}

@article{kynkaanniemi2019improved,
  title={Improved precision and recall metric for assessing generative models},
  author={Kynk{\"a}{\"a}nniemi, Tuomas and Karras, Tero and Laine, Samuli and Lehtinen, Jaakko and Aila, Timo},
  journal=NIPS,
  volume={32},
  year={2019}
}

@article{dhariwal_adm_nips_2021diffusion,
  title={Diffusion models beat gans on image synthesis},
  author={Dhariwal, Prafulla and Nichol, Alexander},
  journal=NIPS,
  volume={34},
  pages={8780--8794},
  year={2021}
}

@inproceedings{deng2009imagenet,
  title={Imagenet: A large-scale hierarchical image database},
  author={Deng, Jia and Dong, Wei and Socher, Richard and Li, Li-Jia and Li, Kai and Fei-Fei, Li},
  booktitle=CVPR,
  pages={248--255},
  year={2009},
  organization={Ieee}
}

@article{li2025minimax,
  title={Minimax-01: Scaling foundation models with lightning attention},
  author={Li, Aonian and Gong, Bangwei and Yang, Bo and Shan, Boji and Liu, Chang and Zhu, Cheng and Zhang, Chunhao and Guo, Congchao and Chen, Da and Li, Dong and others},
  journal={arXiv preprint arXiv:2501.08313},
  year={2025}
}

@article{shen2024scaling,
  title={Scaling laws for linear complexity language models},
  author={Shen, Xuyang and Li, Dong and Leng, Ruitao and Qin, Zhen and Sun, Weigao and Zhong, Yiran},
  journal={arXiv preprint arXiv:2406.16690},
  year={2024}
}

@article{qin_lightnet_2024,
  title={You only scan once: Efficient multi-dimension sequential modeling with lightnet},
  author={Qin, Zhen and Mao, Yuxin and Shen, Xuyang and Li, Dong and Zhang, Jing and Dai, Yuchao and Zhong, Yiran},
  journal={arXiv preprint arXiv:2405.21022},
  year={2024}
}

@article{vaswani2017attention,
  title={Attention is all you need},
  author={Vaswani, Ashish and Shazeer, Noam and Parmar, Niki and Uszkoreit, Jakob and Jones, Llion and Gomez, Aidan N and Kaiser, {\L}ukasz and Polosukhin, Illia},
  journal=NIPS,
  volume={30},
  year={2017}
}

@inproceedings{qin_transnormer_emnlp_2022,
  title={The Devil in Linear Transformer},
  author={Qin, Zhen and Han, Xiaodong and Sun, Weixuan and Li, Dongxu and Kong, Lingpeng and Barnes, Nick and Zhong, Yiran},
  booktitle=EMNLP,
  pages={7025--7041},
  year={2022}
}

@article{2307.14995,
author = {Zhen Qin and Dong Li and Weigao Sun and Weixuan Sun and Xuyang Shen and Xiaodong Han and Yunshen Wei and Baohong Lv and Xiao Luo and Yu Qiao and Yiran Zhong},
title = {TransNormerLLM: A Faster and Better Large Language Model with Improved TransNormer},
Year = {2023},
journal = {arXiv preprint arXiv:2307.14995},
}

@article{qin2024hgrn2,
  title={HGRN2: Gated linear RNNs with state expansion},
  author={Qin, Zhen and Yang, Songlin and Sun, Weixuan and Shen, Xuyang and Li, Dong and Sun, Weigao and Zhong, Yiran},
  journal={arXiv preprint arXiv:2404.07904},
  year={2024}
}

@article{qin_hgrn_nips_2024,
  title={Hierarchically gated recurrent neural network for sequence modeling},
  author={Qin, Zhen and Yang, Songlin and Zhong, Yiran},
  journal=nips,
  volume={36},
  year={2024}
}

@article{sun2023retentive,
  title={Retentive network: A successor to transformer for large language models},
  author={Sun, Yutao and Dong, Li and Huang, Shaohan and Ma, Shuming and Xia, Yuqing and Xue, Jilong and Wang, Jianyong and Wei, Furu},
  journal={arXiv preprint arXiv:2307.08621},
  year={2023}
}

@article{yang_GLA_2023,
  title={Gated linear attention transformers with hardware-efficient training},
  author={Yang, Songlin and Wang, Bailin and Shen, Yikang and Panda, Rameswar and Kim, Yoon},
  journal={arXiv preprint arXiv:2312.06635},
  year={2023}
}

@article{liu2024deepseekv3,
  title={Deepseek-v3 technical report},
  author={Liu, Aixin and Feng, Bei and Xue, Bing and Wang, Bingxuan and Wu, Bochao and Lu, Chengda and Zhao, Chenggang and Deng, Chengqi and Zhang, Chenyu and Ruan, Chong and others},
  journal={arXiv preprint arXiv:2412.19437},
  year={2024}
}

@article{lieber2024jamba,
  title={Jamba: A hybrid transformer-mamba language model},
  author={Lieber, Opher and Lenz, Barak and Bata, Hofit and Cohen, Gal and Osin, Jhonathan and Dalmedigos, Itay and Safahi, Erez and Meirom, Shaked and Belinkov, Yonatan and Shalev-Shwartz, Shai and others},
  journal={arXiv preprint arXiv:2403.19887},
  year={2024}
}

@article{ho_cfg_2022,
  title={Classifier-free diffusion guidance},
  author={Ho, Jonathan and Salimans, Tim},
  journal={arXiv preprint arXiv:2207.12598},
  year={2022}
}

@article{ho_ddpm_nips_2020,
  title={Denoising diffusion probabilistic models},
  author={Ho, Jonathan and Jain, Ajay and Abbeel, Pieter},
  journal=NIPS,
  volume={33},
  pages={6840--6851},
  year={2020}
}

@article{wang2024parallelized,
  title={Parallelized Autoregressive Visual Generation},
  author={Wang, Yuqing and Ren, Shuhuai and Lin, Zhijie and Han, Yujin and Guo, Haoyuan and Yang, Zhenheng and Zou, Difan and Feng, Jiashi and Liu, Xihui},
  journal={arXiv preprint arXiv:2412.15119},
  year={2024}
}

@article{pang2024next,
  title={Next patch prediction for autoregressive visual generation},
  author={Pang, Yatian and Jin, Peng and Yang, Shuo and Lin, Bin and Zhu, Bin and Tang, Zhenyu and Chen, Liuhan and Tay, Francis EH and Lim, Ser-Nam and Yang, Harry and others},
  journal={arXiv preprint arXiv:2412.15321},
  year={2024}
}

@article{li2024scalable,
  title={Scalable autoregressive image generation with mamba},
  author={Li, Haopeng and Yang, Jinyue and Wang, Kexin and Qiu, Xuerui and Chou, Yuhong and Li, Xin and Li, Guoqi},
  journal={arXiv preprint arXiv:2408.12245},
  year={2024}
}

@article{liu2024elucidating,
  title={Elucidating the design space of language models for image generation},
  author={Liu, Xuantong and Hao, Shaozhe and Qi, Xianbiao and Hu, Tianyang and Wang, Jun and Xiao, Rong and Yao, Yuan},
  journal={arXiv preprint arXiv:2410.16257},
  year={2024}
}

@article{li2024imagefolder,
  title={Imagefolder: Autoregressive image generation with folded tokens},
  author={Li, Xiang and Qiu, Kai and Chen, Hao and Kuen, Jason and Gu, Jiuxiang and Raj, Bhiksha and Lin, Zhe},
  journal={arXiv preprint arXiv:2410.01756},
  year={2024}
}

@article{he2024zipar,
  title={Zipar: Accelerating autoregressive image generation through spatial locality},
  author={He, Yefei and Chen, Feng and He, Yuanyu and He, Shaoxuan and Zhou, Hong and Zhang, Kaipeng and Zhuang, Bohan},
  journal={arXiv preprint arXiv:2412.04062},
  year={2024}
}

@misc{yang2024fla,
  title  = {FLA: A Triton-Based Library for Hardware-Efficient Implementations of Linear Attention Mechanism},
  author = {Yang, Songlin and Zhang, Yu},
  url    = {https://github.com/fla-org/flash-linear-attention},
  month  = jan,
  year   = {2024}
}

@article{su2024roformer,
  title={Roformer: Enhanced transformer with rotary position embedding},
  author={Su, Jianlin and Ahmed, Murtadha and Lu, Yu and Pan, Shengfeng and Bo, Wen and Liu, Yunfeng},
  journal={Neurocomputing},
  volume={568},
  pages={127063},
  year={2024},
  publisher={Elsevier}
}

@article{chen2025minimaxM1,
  title={MiniMax-M1: Scaling Test-Time Compute Efficiently with Lightning Attention},
  author={Chen, Aili and Li, Aonian and Gong, Bangwei and Jiang, Binyang and Fei, Bo and Yang, Bo and Shan, Boji and Yu, Changqing and Wang, Chao and Zhu, Cheng and others},
  journal={arXiv preprint arXiv:2506.13585},
  year={2025}
}

@inproceedings{mao2024tavgbench,
  title={TAVGBench: Benchmarking text to audible-video generation},
  author={Mao, Yuxin and Shen, Xuyang and Zhang, Jing and Qin, Zhen and Zhou, Jinxing and Xiang, Mochu and Zhong, Yiran and Dai, Yuchao},
  booktitle=ACMMM,
  pages={6607--6616},
  year={2024}
}

@inproceedings{yu2025randomized,
  title={Randomized autoregressive visual generation},
  author={Yu, Qihang and He, Ju and Deng, Xueqing and Shen, Xiaohui and Chen, Liang-Chieh},
  booktitle=CVPR,
  pages={18431--18441},
  year={2025}
}

@inproceedings{mao2026learning,
  title={Learning spatial decay for vision transformers},
  author={Mao, Yuxin and Qin, Zhen and Zhou, Jinxing and Fan, Bin and Zhang, Jing and Zhong, Yiran and Dai, Yuchao},
  booktitle=AAAI,
  volume={40},
  number={10},
  pages={7945--7953},
  year={2026}
}

@article{mao2026ldt,
  title={Ldt: Efficient scalable video generation using linear diffusion transformer},
  author={Mao, Yuxin and Qin, Zhen and Shen, Xuyang and Zhou, Jinxing and Zhang, Jing and Zhong, Yiran and Dai, Yuchao},
  journal=TMM,
  year={2026},
  publisher={IEEE}
}

@inproceedings{mao2023multimodal,
  title={Multimodal variational auto-encoder based audio-visual segmentation},
  author={Mao, Yuxin and Zhang, Jing and Xiang, Mochu and Zhong, Yiran and Dai, Yuchao},
  booktitle=ICCV,
  pages={954--965},
  year={2023},
  organization={IEEE}
}

@article{mao2024generative,
  title={Generative transformer for accurate and reliable salient object detection},
  author={Mao, Yuxin and Zhang, Jing and Wan, Zhexiong and Tian, Xinyu and Li, Aixuan and Lv, Yunqiu and Dai, Yuchao},
  journal=TCSVT,
  volume={35},
  number={2},
  pages={1041--1054},
  year={2024},
  publisher={IEEE}
}

@article{mao2025contrastive,
  title={Contrastive conditional latent diffusion for audio-visual segmentation},
  author={Mao, Yuxin and Zhang, Jing and Xiang, Mochu and Lv, Yunqiu and Li, Dong and Zhong, Yiran and Dai, Yuchao},
  journal=TIP,
  volume={34},
  pages={4108--4119},
  year={2025},
  publisher={IEEE}
}

@inproceedings{hao2024improving,
  title={Improving audio-visual segmentation with bidirectional generation},
  author={Hao, Dawei and Mao, Yuxin and He, Bowen and Han, Xiaodong and Dai, Yuchao and Zhong, Yiran},
  booktitle=AAAI,
  volume={38},
  number={3},
  pages={2067--2075},
  year={2024}
}

\end{document}